\documentclass[10pt,twocolumn,letterpaper]{article}

\usepackage{cvpr}
\usepackage{times}
\usepackage{epsfig}
\usepackage{graphicx}
\usepackage{amsmath}
\usepackage{amssymb}
\usepackage{soul}

\usepackage[british,english,american]{babel}

\newcommand{\app}{\raise.17ex\hbox{$\scriptstyle\sim$}}
\newcommand{\smaller}[1]{\fontsize{7pt}{1em}\selectfont{#1}}

\newlength\savewidth\newcommand\shline{\noalign{\global\savewidth\arrayrulewidth
  \global\arrayrulewidth 1pt}\hline\noalign{\global\arrayrulewidth\savewidth}}
\newcommand{\tablestyle}[2]{\setlength{\tabcolsep}{#1}\renewcommand{\arraystretch}{#2}\centering\footnotesize}
\makeatletter\renewcommand\paragraph{\@startsection{paragraph}{4}{\z@}
  {.5em \@plus1ex \@minus.2ex}{-.5em}{\normalfont\normalsize\bfseries}}\makeatother

\usepackage{xcolor}
\definecolor{citecolor}{HTML}{0071bc}
\usepackage[pagebackref=false,breaklinks=true,letterpaper=true,colorlinks,citecolor=citecolor,bookmarks=false]{hyperref}

\usepackage{colortbl}
\definecolor{Gray}{gray}{0.5}
\definecolor{GrayBG}{gray}{0.95}

\cvprfinalcopy 

\def\cvprPaperID{****} 

\begin{document}

\title{\vspace{-1em} Improved Baselines with Momentum Contrastive Learning \vspace{-.5em}}

\author{
Xinlei Chen \quad Haoqi Fan \quad Ross Girshick  \quad Kaiming He \vspace{.3em} \\
Facebook AI Research (FAIR)
\vspace{-.5em}
}

\maketitle

\begin{abstract}
\vspace{-.5em}
Contrastive unsupervised learning has recently shown encouraging progress, \eg, in Momentum~Contrast (MoCo) and SimCLR.
In this note, we verify the effectiveness of two of SimCLR's design improvements by implementing them in the MoCo framework.
With simple modifications to MoCo---namely, using an MLP projection head and more data augmentation---we establish stronger baselines that outperform SimCLR and do not require large training batches. We hope this will make state-of-the-art unsupervised learning research more accessible. Code will be made public.

\vspace{-.5em}
\end{abstract}

\section{Introduction}

Recent studies on unsupervised representation learning from images \cite{Wu2018a,Oord2018,Hjelm2019,Ye2019,Bachman2019,Henaff2019,Tian2019,He2019a,Misra2019,Chen2020} are converging on a central concept known as \emph{\mbox{contrastive learning}} \cite{Hadsell2006}. The results are promising: \eg, Momentum Contrast (MoCo) \cite{He2019a} shows that unsupervised pre-training can surpass its ImageNet-supervised counterpart in multiple detection and segmentation tasks, and SimCLR \cite{Chen2020} further reduces the gap in linear classifier performance between \mbox{unsupervised} and supervised pre-training representations.

\begin{figure}[t]
\centering
\includegraphics[width=.9\linewidth]{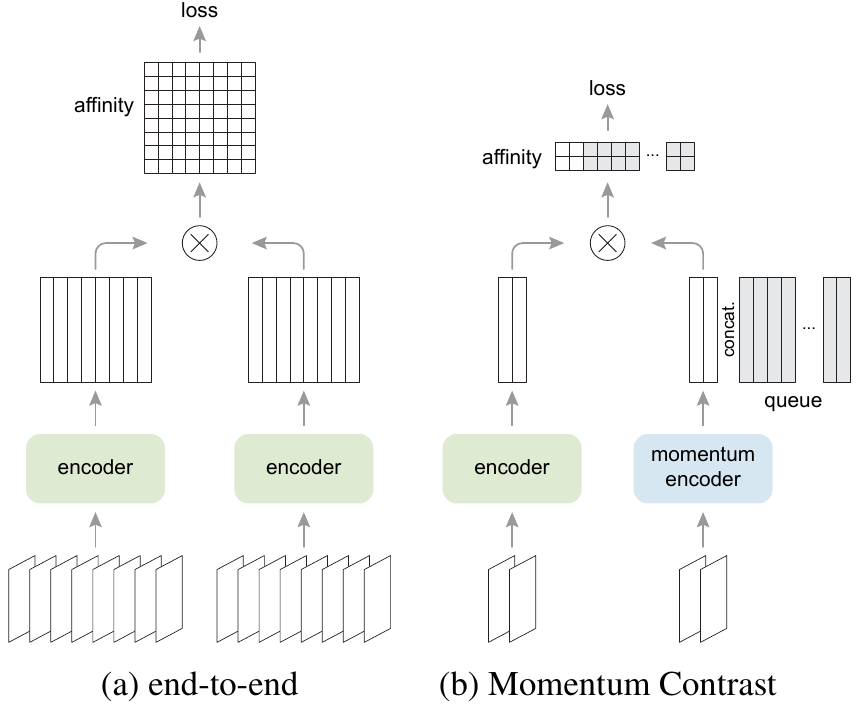}
\caption{A \textbf{batching} perspective of two optimization mechanisms for contrastive learning. Images are encoded into a representation space, in which pairwise affinities are computed.
\label{fig:teaser}
}
\end{figure}

This note establishes stronger and more feasible baselines built in the MoCo framework. We report that two design improvements used in SimCLR, namely, an MLP projection head and stronger data augmentation, are orthogonal to the frameworks of MoCo and SimCLR, and when used with MoCo they lead to better image classification and object detection transfer learning results.
Moreover, the MoCo framework can process a large set of negative samples \mbox{without} requiring large training batches (Fig.~\ref{fig:teaser}). In contrast to SimCLR's large 4k$\app$8k batches, which require TPU support, our ``\textbf{MoCo v2}" baselines can run on a typical 8-GPU machine and achieve better results than SimCLR. 
We hope these improved baselines will provide a reference for future research in unsupervised learning.

\section{Background}

\paragraph{Contrastive learning.}

Contrastive learning \cite{Hadsell2006} is a framework that learns similar/dissimilar representations from data that are organized into similar/dissimilar pairs.
This can be formulated as a dictionary look-up problem. An effective contrastive loss function, called InfoNCE \cite{Oord2018}, is:
\begin{equation}
\small
\mathcal{L}_{q, k^+, \{k^-\}} = -\log \frac{\exp(q{\cdot}k^+ / \tau)}{\exp(q{\cdot}k^+ / \tau) + {\displaystyle\sum_{k^-}}\exp(q{\cdot}k^-  / \tau)}.
\label{eq:infonce}
\end{equation}
Here $q$ is a query representation, $k^+$ is a representation of the positive (similar) key sample, and $\{k^-\}$ are representations of the negative (dissimilar) key samples. $\tau$ is a temperature hyper-parameter.
In the \emph{instance discrimination} pretext task \cite{Wu2018a} (used by MoCo and SimCLR),
a query and a key form a positive pair if they are data-augmented versions of the same image, and otherwise form a negative pair.

The contrastive loss (\ref{eq:infonce}) can be minimized by various mechanisms that differ in how the keys are maintained \cite{He2019a}. In an end-to-end mechanism (Fig.~\ref{fig:teaser}a) \cite{Oord2018,Hjelm2019,Ye2019,Bachman2019,Henaff2019,Chen2020}, the negative keys are from the same batch and updated end-to-end by back-propagation.
SimCLR \cite{Chen2020} is based on this mechanism and requires a large batch to provide a large set of negatives.
In the MoCo mechanism (Fig.~\ref{fig:teaser}b) \cite{He2019a}, the negative keys are maintained in a queue, and only the queries and positive keys are encoded in each training batch. A momentum encoder is adopted to improve the representation consistency between the current and earlier keys. MoCo decouples the batch size from the number of negatives.

\paragraph{Improved designs.}
SimCLR \cite{Chen2020} improves the end-to-end variant of instance discrimination in three aspects: (i) a substantially larger batch (4k or 8k) that can provide more negative samples; (ii) replacing the output \emph{fc} projection head \cite{Wu2018a} with an MLP head; (iii) stronger data augmentation.

In the MoCo framework, a large number of negative samples are readily available; the MLP head and data augmentation are orthogonal to how contrastive learning is instantiated.
Next we study these improvements in MoCo.

\begin{table}[t]
\vspace{-1em}
\begin{center}
\tablestyle{3pt}{1.0}
\begin{tabular}{c|cccc|c|ccc}
~  & \multicolumn{4}{c|}{\smaller{unsup. pre-train}} & \smaller{ImageNet} & \multicolumn{3}{c}{\smaller{VOC detection}} \\
case &
MLP &
aug+ &
cos &
epochs & 
\multicolumn{1}{c|}{acc.} &
AP$_\text{50}$ & AP & AP$_\text{75}$ \\
\shline
\color{Gray}{supervised} & & & & &
\color{Gray}{76.5} &
\color{Gray}{81.3} &
\color{Gray}{53.5} &
\color{Gray}{58.8} \\
MoCo v1 & & & & 200 & 60.6 & 81.5 & 55.9 & 62.6 \\
(a) & \checkmark & & & 200 & 66.2 & 82.0 & 56.4 & 62.6 \\
(b) & & \checkmark & & 200 & 63.4 & 82.2 & 56.8 & 63.2 \\
(c) & \checkmark & \checkmark & & 200 & 67.3 & \textbf{82.5} & 57.2 & 63.9 \\
(d) & \checkmark & \checkmark & \checkmark & 200 & 67.5 & 82.4 & 57.0 & 63.6 \\
(e) & \checkmark & \checkmark & \checkmark & \textbf{800} & \textbf{71.1} & \textbf{82.5} & \textbf{57.4} & \textbf{64.0} \\
\end{tabular}
\end{center}
\vspace{-.5em}
\caption{\textbf{Ablation of MoCo baselines}, evaluated by ResNet-50 for (i) ImageNet linear classification, and (ii) fine-tuning VOC object detection (mean of 5 trials).
``\textbf{MLP}'': with an MLP head;
``\textbf{aug+}'': with extra blur augmentation;
``\textbf{cos}'': cosine learning rate schedule.
}
\label{tab:ablation}
\vspace{-1em}
\end{table}

\section{Experiments}

\paragraph{Settings.}
Unsupervised learning is conducted on the 1.28M \mbox{ImageNet} \cite{Deng2009} training set.
We follow two common protocols for evaluation.
(i) \emph{ImageNet linear classification}: features are frozen and a supervised linear classifier is trained; we report 1-crop (224$\times$224), top-1 validation accuracy. (ii) \emph{Transferring to VOC object detection} \cite{Everingham2010}: a Faster R-CNN detector \cite{Ren2015} (C4-backbone) is fine-tuned end-to-end on the VOC 07+12 \texttt{trainval} set\footnote{For all entries (including the supervised and MoCo v1 baselines), we fine-tune for 24k iterations on VOC, up from 18k in \cite{He2019a}.} and evaluated on the VOC 07 \texttt{test} set using the COCO suite of metrics \cite{Lin2014}. We use the same hyper-parameters (except when noted) and codebase as MoCo \cite{He2019a}.
All results use a standard-size \mbox{ResNet-50} \cite{He2016}.

\paragraph{MLP head.} Following \cite{Chen2020}, we replace the \emph{fc} head in MoCo with a 2-layer MLP head (hidden layer 2048-d, with ReLU). Note this only influences the unsupervised training stage; the \emph{linear} classification or transferring stage does not use this MLP head.
Also, following \cite{Chen2020}, we search for an optimal $\tau$ \wrt ImageNet linear classification accuracy:
\begin{center}
\vspace{-.3em}
\tablestyle{8pt}{1.0}	
\begin{tabular}{ccccccc}
$\tau$ & 0.07 & 0.1 & 0.2 & 0.3 & 0.4 & 0.5 \\
\shline
w/o MLP & 60.6 & \textbf{60.7} & 59.0 & 58.2 & 57.2 & 56.4 \\
w/ MLP & 62.9 & 64.9 & \textbf{66.2} & 65.7 & 65.0 & 64.3 \\
\end{tabular}
\vspace{-.3em}
\end{center}
Using the default $\tau =$ 0.07 \cite{Wu2018a,He2019a}, pre-training with the MLP head improves from 60.6\% to 62.9\%; switching to the optimal value for MLP (0.2), the accuracy increases to 66.2\%. Table~\ref{tab:ablation}(a) shows its detection results: in contrast to the big leap on ImageNet, the detection gains are smaller.

\paragraph{Augmentation.} We extend the original augmentation in \cite{He2019a} by including the blur augmentation in \cite{Chen2020} (we find the stronger color distortion in \cite{Chen2020} has diminishing gains in our higher baselines).
The extra augmentation alone (\ie, no MLP) improves the MoCo baseline on ImageNet by 2.8\% to 63.4\%, Table~\ref{tab:ablation}(b).
Interestingly, its detection accuracy is higher than that of using the MLP alone, Table~\ref{tab:ablation}(b) \vs (a), despite much lower linear classification accuracy (63.4\% \vs 66.2\%). This indicates that {\emph{linear classification accuracy is not monotonically related to transfer performance in detection}}.
With the MLP, the extra augmentation boosts ImageNet accuracy to 67.3\%, Table~\ref{tab:ablation}(c). 

\paragraph{Comparison with SimCLR.} Table~\ref{tab:main} compares SimCLR \cite{Chen2020} with
our results, referred to as MoCo~v2.
For fair comparisons, we also study a cosine (half-period) learning rate schedule \cite{Loshchilov2016} which SimCLR adopts. See Table~\ref{tab:ablation}\mbox{(d, e)}.
Using pre-training with 200 epochs and a batch size of 256, MoCo v2 achieves 67.5\% accuracy on ImageNet: this is 5.6\% higher than SimCLR \emph{under the same epochs and batch size}, and better than SimCLR's large-batch result 66.6\%. With 800-epoch pre-training, \mbox{MoCo~v2} achieves 71.1\%, outperforming SimCLR's 69.3\% with 1000 epochs.

\begin{table}[t]
\vspace{-1em}
\begin{center}
\tablestyle{6pt}{1.0}
\begin{tabular}{l|cccrr|c}
~  & \multicolumn{5}{c|}{\smaller{unsup. pre-train}} & \smaller{ImageNet} \\
case &
MLP &
aug+ &
cos & 
epochs &
batch &
acc. \\
\shline
MoCo v1 \cite{He2019a} & & & & 200 & 256 & 60.6 \\
SimCLR \cite{Chen2020} & \checkmark & \checkmark & \checkmark & 200 & 256 & 61.9 \\
SimCLR \cite{Chen2020} & \checkmark & \checkmark & \checkmark & 200 & 8192 & 66.6 \\
\rowcolor{GrayBG} \textbf{MoCo v2} & \checkmark & \checkmark & \checkmark & 200 & 256 & \textbf{67.5}  \\
\hline
\multicolumn{6}{l}{\emph{results of \textbf{longer} unsupervised training follow:}} \\
\hline
SimCLR \cite{Chen2020} & \checkmark & \checkmark & \checkmark & 1000 & 4096 & 69.3 \\
\rowcolor{GrayBG} \textbf{MoCo v2} & \checkmark & \checkmark & \checkmark & 800 & 256 & \textbf{71.1} \\
\end{tabular}
\end{center}
\vspace{-.5em}
\caption{\textbf{MoCo \vs SimCLR}: ImageNet linear classifier accuracy (\textbf{ResNet-50, \mbox{1-crop 224$\times$224}}), trained on features from unsupervised pre-training.
``aug+'' in SimCLR includes blur and stronger color distortion.
SimCLR ablations are from \mbox{Fig.~9} in \cite{Chen2020} (we thank the authors for providing the numerical results).
}
\label{tab:main}
\end{table}

\begin{table}[t]
\begin{center}
\tablestyle{6pt}{1.0}
\begin{tabular}{cccc}
mechanism & batch & memory / GPU & time / 200-ep. \\
\shline
\rowcolor{GrayBG} MoCo & 256 & \textbf{5.0G} & \textbf{53 hrs} \\
end-to-end & 256 & 7.4G & 65 hrs \\
end-to-end & 4096 & 93.0G$^\dagger$ & n/a \\
\end{tabular}
\end{center}
\vspace{-.5em}
\caption{\textbf{Memory and time cost} in 8 V100 16G GPUs, implemented in PyTorch.
$^\dagger$: based on our estimation.
}
\label{tab:cost}
\vspace{-.5em}
\end{table}

\paragraph{Computational cost.}
In Table~\ref{tab:cost} we report the memory and time cost of our implementation.
The end-to-end case reflects the SimCLR cost in GPUs (instead of TPUs in \cite{Chen2020}). 
The 4k batch size is intractable even in a high-end 8-GPU machine.
Also, under the same batch size of 256, the end-to-end variant is still more costly in memory and time, because it back-propagates to both $q$ and $k$ encoders, while MoCo back-propagates to the $q$ encoder only.

\vspace{1em}
Table~\ref{tab:main} and \ref{tab:cost} suggest that large batches are not necessary for good accuracy, and state-of-the-art results can be made more accessible.
The improvements we investigate require only a few lines of code changes to MoCo~v1, and we will make the code public to facilitate future research.

{\small
\bibliographystyle{ieee_fullname}
\bibliography{mocov2}

\begin{thebibliography}{10}\itemsep=-1pt

\bibitem{Bachman2019}
Philip Bachman, R~Devon Hjelm, and William Buchwalter.
\newblock Learning representations by maximizing mutual information across
  views.
\newblock {\em arXiv:1906.00910}, 2019.

\bibitem{Chen2020}
Ting Chen, Simon Kornblith, Mohammad Norouzi, and Geoffrey Hinton.
\newblock A simple framework for contrastive learning of visual
  representations.
\newblock {\em arXiv:2002.05709}, 2020.

\bibitem{Deng2009}
Jia Deng, Wei Dong, Richard Socher, Li-Jia Li, Kai Li, and Li Fei-Fei.
\newblock {ImageNet: A large-scale hierarchical image database}.
\newblock In {\em CVPR}, 2009.

\bibitem{Everingham2010}
Mark Everingham, Luc Van~Gool, Christopher~KI Williams, John Winn, and Andrew
  Zisserman.
\newblock {The {PASCAL} Visual Object Classes (VOC) Challenge}.
\newblock {\em IJCV}, 2010.

\bibitem{Hadsell2006}
Raia Hadsell, Sumit Chopra, and Yann LeCun.
\newblock Dimensionality reduction by learning an invariant mapping.
\newblock In {\em CVPR}, 2006.

\bibitem{He2019a}
Kaiming He, Haoqi Fan, Yuxin Wu, Saining Xie, and Ross Girshick.
\newblock Momentum contrast for unsupervised visual representation learning.
\newblock {\em arXiv:1911.05722}, 2019.

\bibitem{He2016}
Kaiming He, Xiangyu Zhang, Shaoqing Ren, and Jian Sun.
\newblock Deep residual learning for image recognition.
\newblock In {\em CVPR}, 2016.

\bibitem{Hjelm2019}
R~Devon Hjelm, Alex Fedorov, Samuel Lavoie-Marchildon, Karan Grewal, Adam
  Trischler, and Yoshua Bengio.
\newblock Learning deep representations by mutual information estimation and
  maximization.
\newblock In {\em ICLR}, 2019.

\bibitem{Henaff2019}
Olivier~J. Hénaff, Aravind Srinivas, Jeffrey~De Fauw, Ali Razavi, Carl
  Doersch, S.~M.~Ali Eslami, and Aaron van~den Oord.
\newblock Data-efficient image recognition with contrastive predictive coding.
\newblock {\em arXiv:1905.09272v2}, 2019.

\bibitem{Lin2014}
Tsung-Yi Lin, Michael Maire, Serge Belongie, James Hays, Pietro Perona, Deva
  Ramanan, Piotr Doll{\'a}r, and C~Lawrence Zitnick.
\newblock {Microsoft COCO: Common objects in context}.
\newblock In {\em ECCV}. 2014.

\bibitem{Loshchilov2016}
Ilya Loshchilov and Frank Hutter.
\newblock {SGDR}: Stochastic gradient descent with warm restarts.
\newblock In {\em ICLR}, 2017.

\bibitem{Misra2019}
Ishan Misra and Laurens van~der Maaten.
\newblock Self-supervised learning of pretext-invariant representations.
\newblock {\em arXiv:1912.01991}, 2019.

\bibitem{Oord2018}
Aaron van~den Oord, Yazhe Li, and Oriol Vinyals.
\newblock Representation learning with contrastive predictive coding.
\newblock {\em arXiv:1807.03748}, 2018.

\bibitem{Ren2015}
Shaoqing Ren, Kaiming He, Ross Girshick, and Jian Sun.
\newblock {Faster R-CNN}: Towards real-time object detection with region
  proposal networks.
\newblock In {\em NeurIPS}, 2015.

\bibitem{Tian2019}
Yonglong Tian, Dilip Krishnan, and Phillip Isola.
\newblock Contrastive multiview coding.
\newblock {\em arXiv:1906.05849}, 2019.

\bibitem{Wu2018a}
Zhirong Wu, Yuanjun Xiong, Stella Yu, and Dahua Lin.
\newblock Unsupervised feature learning via non-parametric instance
  discrimination.
\newblock In {\em CVPR}, 2018.

\bibitem{Ye2019}
Mang Ye, Xu Zhang, Pong~C Yuen, and Shih-Fu Chang.
\newblock Unsupervised embedding learning via invariant and spreading instance
  feature.
\newblock In {\em CVPR}, 2019.

\end{thebibliography}
}

\end{document}